\documentclass[sigconf]{acmart}

\usepackage{subfigure}

\AtBeginDocument{%
  \providecommand\BibTeX{{%
    \normalfont B\kern-0.5em{\scshape i\kern-0.25em b}\kern-0.8em\TeX}}}

\setcopyright{acmcopyright}
\copyrightyear{2020}
\acmYear{2020}
\acmDOI{}

\acmConference[MM '20]{MM '20: the 28th ACM International Conference on Multimedia}{October 12--16, 2020}{Seattle, USA}

\acmSubmissionID{291}

\graphicspath{{figures/}}

\begin{document}

\title{Dynamic GCN: Context-enriched Topology Learning for Skeleton-based Action Recognition}


\author{Fanfan Ye$^{1,2}$, Shiliang Pu$^{1\ast}$, Qiaoyong Zhong$^1$, Chao Li$^1$, Di Xie$^1$, Huiming Tang$^2$}
\thanks{$^\ast$Corresponding author.}

\affiliation{\institution{$^1$ Hikvision Research Institute}}
\email{{yefanfan,pushiliang.hri,zhongqiaoyong,lichao15,xiedi}@hikvision.com}

\affiliation{\institution{$^2$ College of Information Science \& Electronic Engineering, Zhejiang University}}
\email{thm@zju.edu.cn}

\renewcommand{\shortauthors}{Ye, Pu and Zhong, et al.}
\fancyhead{}  

\begin{abstract}
 Graph Convolutional Networks (GCNs) have attracted increasing interests for the task of skeleton-based action recognition. The key lies in the design of the graph structure, which encodes skeleton topology information. In this paper, we propose Dynamic GCN, in which a novel convolutional neural network named Context-encoding Network (CeN) is introduced to learn skeleton topology automatically. In particular, when learning the dependency between two joints, contextual features from the rest joints are incorporated in a global manner. CeN is extremely lightweight yet effective, and can be embedded into a graph convolutional layer. By stacking multiple CeN-enabled graph convolutional layers, we build Dynamic GCN. Notably, as a merit of CeN, dynamic graph topologies are constructed for different input samples as well as graph convolutional layers of various depths. Besides, three alternative context modeling architectures are well explored, which may serve as a guideline for future research on graph topology learning. CeN brings only \textasciitilde7\% extra FLOPs for the baseline model, and Dynamic GCN achieves better performance with $2\times$\textasciitilde$4\times$ fewer FLOPs than existing methods. By further combining static physical body connections and motion modalities, we achieve state-of-the-art performance on three large-scale benchmarks, namely NTU-RGB+D, NTU-RGB+D 120 and Skeleton-Kinetics.
\end{abstract}




\maketitle

\section{Introduction}
Human action recognition is a popular topic in the area of computer vision. Especially, skeleton-based action recognition has attracted more and more attention. Compared with RGB data, skeleton data are considered as a more robust representation for human action dynamics. Meanwhile, skeleton data are extremely compact in terms of data size. This makes it possible to design more lightweight models. Skeleton data can be easily captured by depth cameras (e.g. Kinetics) or estimated with human pose estimation algorithms~\cite{chen2018cascaded,wei2016convolutional,cao2018openpose}.

\begin{figure}[t]
\centering
\includegraphics[width=\columnwidth]{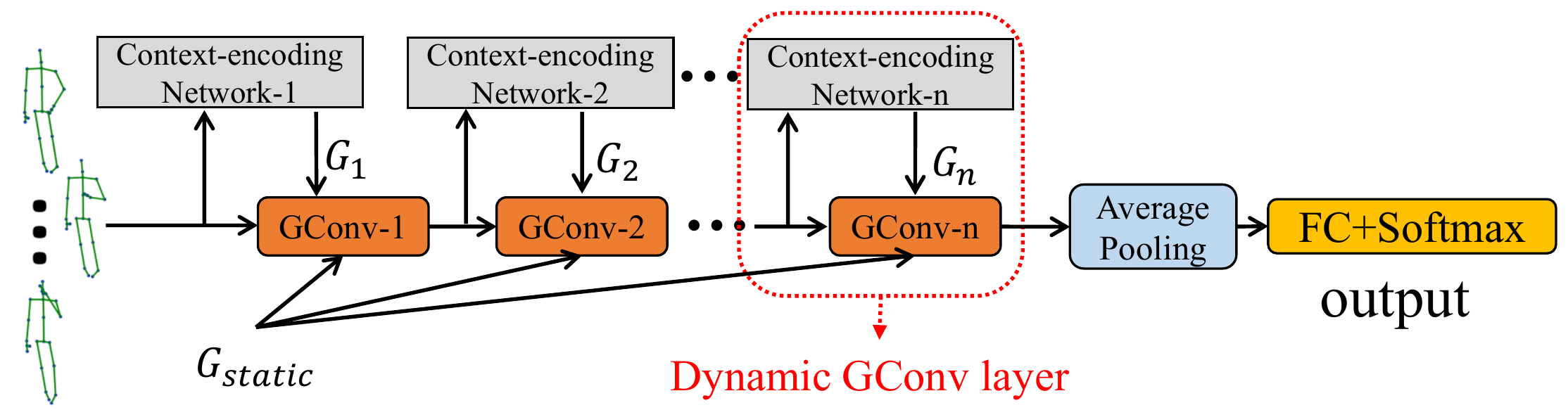}
\caption{An overview of the proposed Dynamic GCN framework. The backbone is comprised of $n$ Dynamic GConv layers, which accept two graph topologies as input, i.e. static topology $G_{static}$ of pre-defined physical connections and dynamic topology $\left\{ G_1, G_2, \dots, G_n \right\}$ predicted by the context-encoding networks.}
\label{fig1}
\end{figure}

In the task of skeleton-based action recognition, we are given a time series of human joint coordinates and expected to predict the action being performed. Considering the sequence property, Recurrent Neural Networks (RNNs) are a natural choice and have been widely studied~\cite{shahroudy,zhang:J,zhang2018adding,liu2016spatio,zhang2017view}. On the other hand, some recent works attempted to cast a skeleton sequence as a pseudo-image. Then Convolutional Neural Networks (CNNs) are employed to classify the image directly~\cite{li2017skeleton,lili}, which also achieve great success. The advantage of deep neural networks (DNNs) lies in their powerful feature learning capability. Nevertheless, none of these methods explicitly exploits the skeleton topology information, which is very informative for discriminating different action categories. \cite{yanspatial} first introduced Graph Convolutional Networks (GCNs)~\cite{kipf2017} in the context of skeleton-based action recognition. They model skeleton data as a graph and extract the topology information by an adjacency matrix according to the physical connections of human body. However, the dependencies among skeleton joints of different samples vary, especially when they are performing different actions. Such topology information derived from fixed graph is relatively weak. Recently, there have been some attempts~\cite{shi2019two,zhang2019semantics,li2019spatio,liu2020} to construct different graphs for different samples. They are basically inspired by the non-local operation~\cite{buades2005non}, where a distance metric like inner product is utilized to measure the degree of dependency between two arbitrary skeleton joints. To some extent, the topology information is strengthened and the recognition performance gets improved. However, they also bring three issues for future improvement. 1)~Non-local-based methods, measuring the dependency between two skeleton joints while ignoring the influence of all other contextual joints, is essentially a local method. We argue that besides the underlying two joints, the contextual information from the rest joints is critical for learning reliable and stable topology. 2)~Using an arbitrary function like inner product to compute the dependency between two joints introduces strong prior knowledge, which may not be optimal. 3)~In the skeleton dynamic system, non-local-based methods consider the dependency of every pair of joints undirected. Since the contextual information of each joint is different, the dependency should be directed. Moreover, for different pairs of queries, their similarities yielded by the non-local-based methods may be almost identical~\cite{cao2019gcnet}.

In this work, we propose a hybrid GCN-CNN framework named Dynamic GCN as shown in Figure~\ref{fig1}. We aim to attack the weaknesses of existing learning-based skeleton topology by leveraging the feature learning capability of CNNs. Specifically, a novel convolutional neural network named Context-encoding Network (CeN) is introduced to learn skeleton topology automatically. It can be embedded into a graph convolutional (GConv) layer and learned end-to-end. Compared with non-local-based methods, CeN has the following advantages. 1)~CeN takes full account of the contextual information of each joint from a global perspective. 2)~CeN is completely data-driven and does not require any prior knowledge, which is more flexible. 3)~CeN regards the dependency of each pair of joints as directed and yields directed graphs (asymmetric adjacency matrices), which can represent the dynamics of the skeleton system more accurately. 4)~Compared with other topology-learning algorithms, CeN is extremely lightweight yet effective, and can be integrated into the GCN-based methods easily.

Notably, CeN predicts a unique graph topology per-sample as well as per-GConv layer. This feature results in a dynamic graph topology rather than static topology, which enhances the capacity and expressiveness of the model.


For context modeling in CeN, various feature aggregation architectures are explored. As pointed out by~\cite{lili}, for a two-dimensional (2D) convolutional layer, features are aggregated globally along the channel dimension and locally along the spatial dimensions. A skeleton sequence can be represented as a tensor of $C\times T \times N$, where $C$, $T$ and $N$ denote the feature, temporal and joint dimensions respectively. In the context of topology learning, we argue that the contextual information from the surrounding joints are the most important. To this end, in the proposed CeN features are aggregated globally along the joint dimension by treating it as channel. Ablation studies show that it is superior over two other alternatives, where either the temporal or feature dimension is treated as channel. Further discussion is given, which may guide future research on graph topology learning.

In terms of performance, CeN alone surpasses existing non-local-based topology learning methods significantly. Combining dynamic topology predicted by CeN with static topology leads to further performance gain, which indicates that CeN is complementary to static topology. In terms of efficiency, CeN only brings \textasciitilde7\% extra FLOPs for the GCN-based baseline method. By using the joint-level feature aggregation mechanism, Dynamic GCN is $2\times$\textasciitilde$4\times$ less expensive than other GCN-based methods on calculation.

Moreover, by incorporating the spatial and motion modalities of skeleton sequences, our final model achieves state-of-the-art performance on three large-scale benchmarks, namely NTU-RGB+D, NTU-RGB+D 120 and Skeleton-Kinetics.

Our main contributions can be summarized as follows.
\begin{itemize}
\item We propose the Dynamic GCN framework, which leverages the complementary benefits of GCN's topology learning and CNN's feature learning capabilities.
\item We introduce a lightweight context-encoding network, which learns context-enriched dynamic skeleton topology in a global way.
\item Three alternative context modeling architectures are well explored, which may serve as a guideline for future research on graph topology learning.
\item Our final model achieves state-of-the-art performance on three large-scale benchmarks for skeleton-based action recognition.
\end{itemize}

\section{Related Works}

\begin{figure}[t]
\centering
\includegraphics[width=0.8\columnwidth]{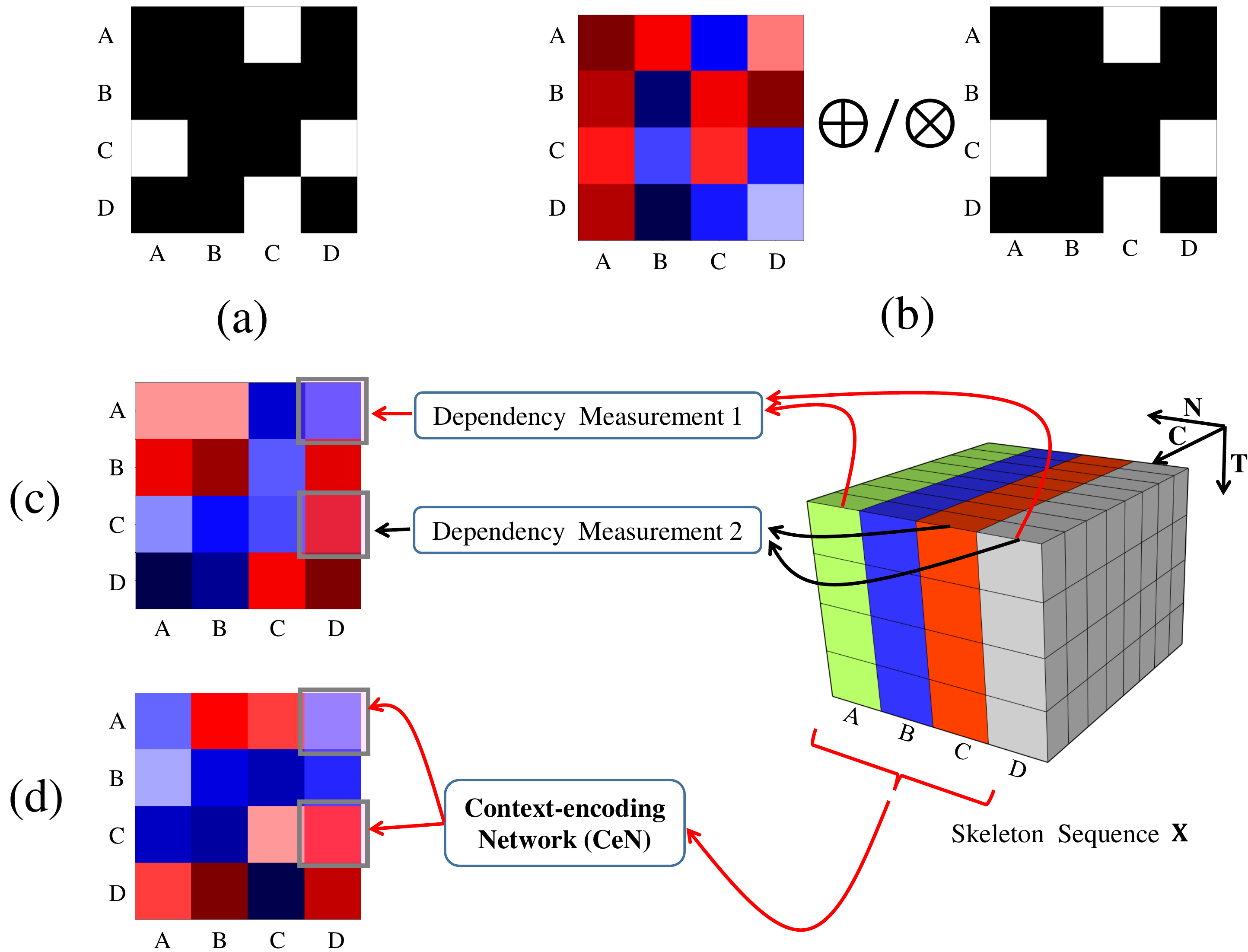} 
  \caption{Comparison of various ways to construct the adjacency matrix. (a) Pre-defined physical adjacency matrix. (b) Learnable mask multiplied or added with physical adjacency matrix. (c) Existing non-local-based adjacency matrix taking only the underlying two joints into account. (d) The proposed dynamic topology predicted by CeN, in which features of all contextual joints are incorporated.}
\label{fig2}
\end{figure}

Skeleton-based action recognition has been dominated by deep learning-based methods. They have been proven more effective than those methods based on hand-crafted features~\cite{vemulapalli2014human,wang2012mining,xia2012view}. We will summarize recent works into two major categories, i.e. DNN-based methods and GCN-based methods.

\subsection{DNN-based Methods}

RNN is a straightforward model for sequence data. \cite{shahroudy} divides the cell of LSTM into five part-cells which correspond to five body parts respectively. \cite{zhang2017view} proposed a view adaptation scheme to automatically regulate observation viewpoints during the occurrence of an action. While RNNs aggregate temporal information sequentially, CNNs are able to encode spatiotemporal information jointly. \cite{li2017skeleton} treats a skeleton sequence as a pseudo-image, in which the coordinate of each skeleton joint $(x, y, z)$ is regarded as three channels of the image. Then a CNN is designed to directly classify the image into action categories. \cite{lili} proposes a co-occurrence feature learning framework, which inspires the global contextual feature aggregation of the proposed CeN. The success of CNNs is attributed to their strong capability in feature representation. However, after converting the irregularly structured skeleton data into regularly structured images, the skeleton topology information is lost. Although CeN is also a CNN model, it is designed for learning of the graph topology rather than final action recognition.

\subsection{GCN-based Methods}
GCN is able to effectively deal with irregularly structured graphs like skeleton data. Given skeleton data with $N$ joints, the graph topology can be well represented by an $N \times N$ adjacency matrix $\mathbf{A}$. The key of GCN-based methods lies in the design of graph topology, i.e. $\mathbf{A}$. The most straightforward way is to define a fixed graph according to the physical connections of human body (Figure~\ref{fig2}(a)), which is adopted in ST-GCN~\cite{yanspatial}. In order to put slight attention on edges, they also create a learnable mask which is multiplied or added with the physical adjacency matrix (Figure~\ref{fig2}(b)). Later, \cite{tang2018deep} adopted the conception of virtual connection as a supplement to physical adjacency matrix.

In the above methods, the adjacency matrices are either pre-defined or fixed after training finishes. To make the graph topology more flexible, \cite{shi2019two,liu2020} and \cite{li2019spatio} attempted to construct different graphs for different samples. Specifically, non-local-based operations are employed to infer the connectedness between two arbitrary joints. As shown in Figure~\ref{fig2}(c), when measuring the dependency between two joints, only features of the underlying two joints are taken into account, while the influence of the contextual joints is ignored. On the contrary, in our Dynamic GCN, features of all contextual joints are fully incorporated with the proposed CeN (Figure~\ref{fig2}(d)). The graph learned in this way can be more robust and expressive.

\section{Method}
In this section, we first briefly recap GCN in the context of skeleton-based action recognition. Then we illustrate the details of the proposed Dynamic GCN framework.

\subsection{GCN for Skeleton-based Action Recognition}

In the context of skeleton-based action recognition, a GCN is typically composed of graph convolutional blocks (GC-blocks) and temporal convolutional blocks (TC-blocks). Given a skeleton sequence of $C \times T \times N$, GC-blocks and TC-blocks aggregate features along the joint ($N$) and temporal ($T$) dimensions respectively. Note that the number of joints $N$ is kept unchanged, while $C$ and $T$ normally differ in different GConv layers. Specifically, GC-blocks can be formulated as:
\begin{equation}
 {{\mathbf{Y}}} = \sum\limits_{k = 1}^K {{\mathbf{\Lambda }}_k^{ - \frac{1}{2}}{{\mathbf{A}}_{k}}} {\mathbf{\Lambda }}_k^{ - \frac{1}{2}}{\mathbf{XW}},
 \label{eq1}
 \end{equation}
where $K$ denotes the number of spatial configurations according to ST-GCN~\cite{yanspatial}. $\mathbf{X}$ and $\mathbf{Y}$ denote the input and output feature maps respectively. ${\mathbf{W}}$ denotes the learnable kernels. For each spatial configuration, ${\mathbf{A}}$ is the adjacency matrix and ${\mathbf{\Lambda}}$ is the diagonal node degree matrix for normalization. Specifically, the degree of node $i$ is computed by $\Lambda^{ii} = \sum\nolimits_j {A^{ij}} + \alpha $, where ${A^{ij}}$ denotes the element of the $i$-th row and $j$-th column in $\mathbf{A}$, and $\alpha$ is added to avoid the all-zero problem.

TC-blocks are normal convolutional layers with a kernel size of $t \times 1$. To learn spatiotemporal features jointly, a GCN is typically built by stacking GC-blocks and TC-blocks alternately.

\subsection{Dynamic GCN}

In this section, we elaborately introduce the proposed Dynamic GCN framework. We first introduce the architecture of Context-encoding Network (CeN) for topology learning. Then we describe how to integrate CeN into GConv layers and build the complete framework.

\subsubsection{Context-encoding Network}

\begin{figure}[t]
\centering
\includegraphics[width=0.5\columnwidth]{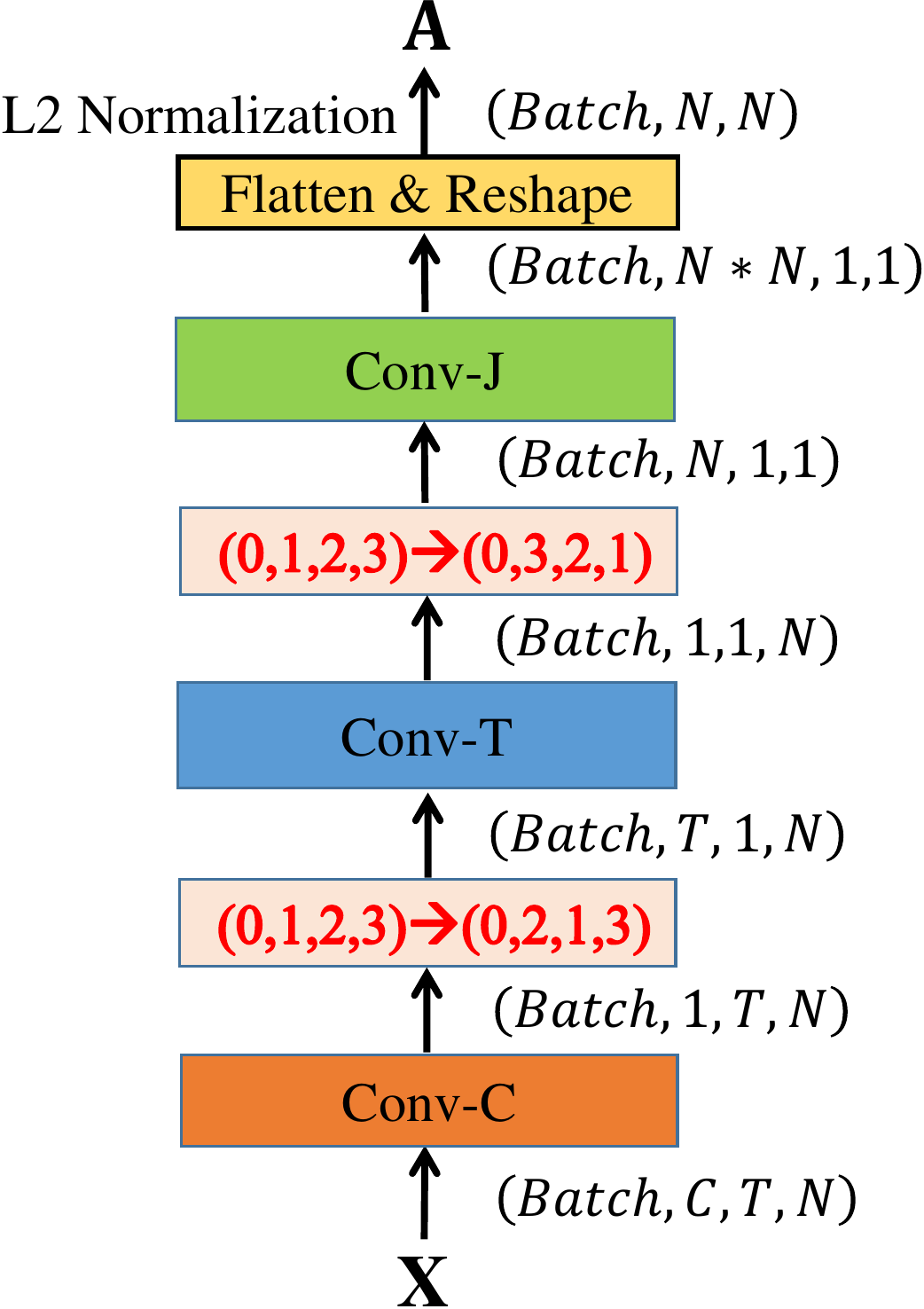} 
  \caption{Architecture of the proposed CeN. It only consists of three 1$\times$1 convolutional layers. Conv-C and Conv-T are first adopted to squeeze the feature and temporal dimensions. And then the joint dimension is treated as channel to acquire the global topology. Feature map permutation (block in pink) is applied on demand. Batch Normalization and ReLU activation function are applied after each convolutional layer.}
\label{fig3}
\end{figure}

The adjacency matrix ${\mathbf{A}}$ in GCN fully represents the graph topology, which corresponds to the dependencies among different skeleton joints. When ${\mathbf{A}}$ is pre-defined with prior knowledge, the topology information is static and limited. Existing learning-based methods~\cite{zhang2019semantics,shi2019two,li2019spatio,liu2020} predict the dependency between two joints $A^{ij}$ independently, and hand-crafted functions (e.g. inner product) to map input features into dependency are employed. On the contrary, we design an extremely lightweight convolutional neural network which takes the whole feature map as input and predicts the full adjacency matrix $\mathbf{A}$ directly. Notably, contextual information along the joint, temporal and feature dimensions are well explored, yielding a more flexible and expressive graph topology.

The architecture of CeN is shown in Figure~\ref{fig3}. Given a feature map of intermediate layer ${\mathbf{X}} \in {R^{C \times T \times N}}$, firstly the feature and temporal dimensions of each joint are squeezed by two 1$\times$1 convolutional layers named Conv-C and Conv-T. Then the joint dimension is treated as the channel of convolution, and a single 1$\times$1 convolutional layer is utilized to map the $N$-dimensional vector into the $N\times N$ adjacency matrix. This design fully considers the impact of all other joints when measuring the dependency between every pair of joints. After that, the topology is represented as matrices with the shape of $\left( {Batch, N, N} \right)$. In addition, the L2 normalization is applied to each row of the adjacency matrices, which eases the optimization.

It is worth noting that CeN learns a dynamic and unique adjacency matrix per-sample as well as per-GConv layer. The graph topology is not shared among different samples even if they belong to the same action class. Rather than hand-crafted functions, the parameters in CeN are learned in a data-driven way without any prior assumption. Moreover, by treating the joint dimension as channel, global contextual information of all joints can be encoded by the trainable kernels.

The adjacency matrix predicted by CeN is fed into a GConv layer as its graph topology representation. With the help of L2 normalization, the normalization of node degree in Eq.~\eqref{eq1} is unnecessary. For simplicity, henceforth we use $\mathbf{G}$ to refer to either normalized static adjacency matrix or learned adjacency matrix.

\subsubsection{Dynamic GConv Layer}
\begin{figure}[t]
\centering
\includegraphics[width=0.78\columnwidth]{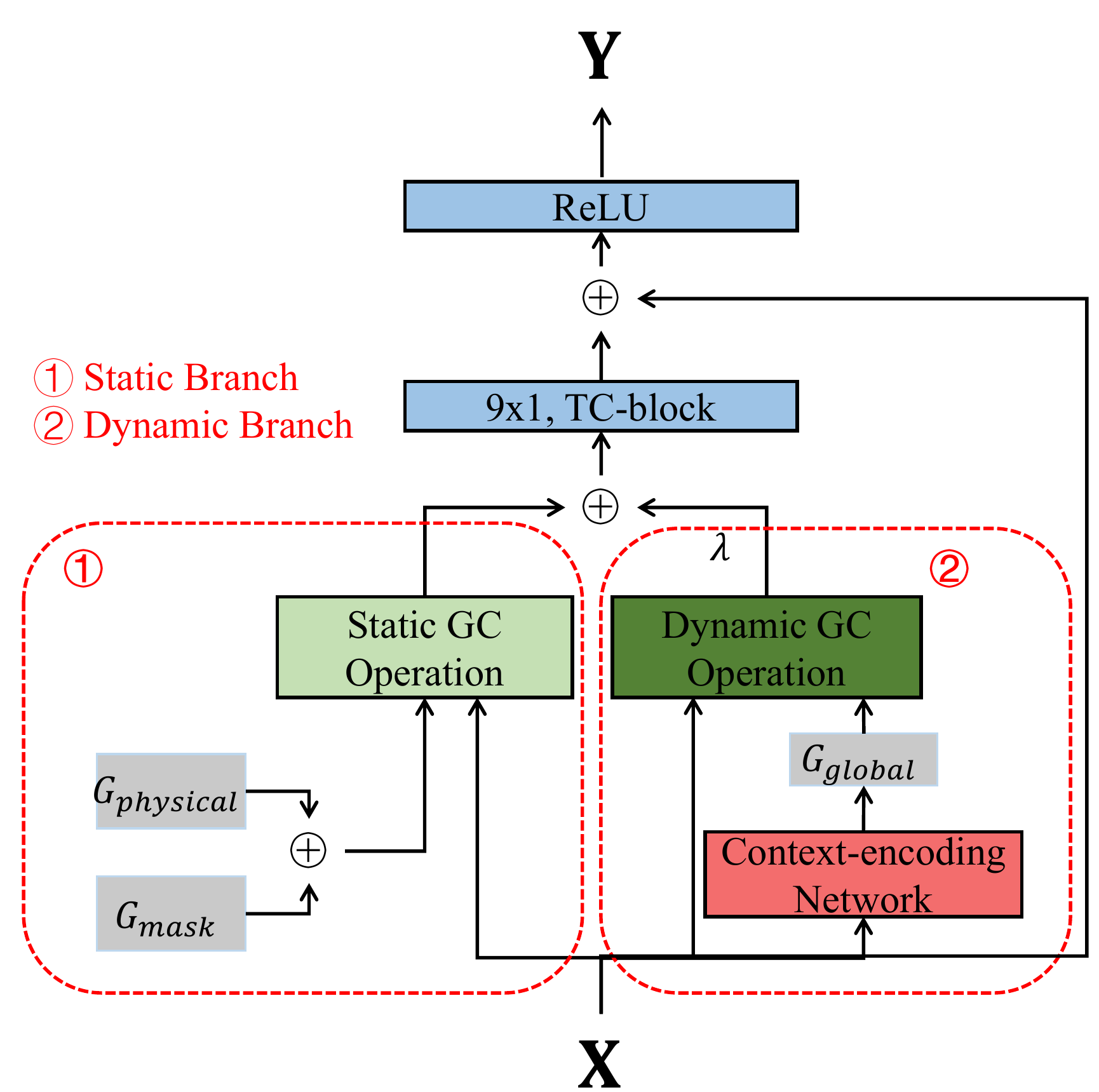} 
  \caption{Pipeline of the Dynamic GConv layer. Topology features derived from static graph (static branch) and graph predicted by CeN (dynamic branch) are fused. Afterwards, a TC-block is appended.}
\label{fig4}
\end{figure}

Figure~\ref{fig4} shows the pipeline of the Dynamic GConv layer, which is the basic building block of the Dynamic GCN framework. Besides the dynamic graph predicted by CeN, static graphs are also incorporated. The static branch takes the skeleton features from the last layer and the physical graph ${{\mathbf{G}}_{physical}}$ with learnable parameter-mask ${{\mathbf{G}}_{mask}}$ as input. The dynamic branch takes the skeleton features and the context-enriched graph ${{\mathbf{G}}_{global}}$ as input. Their outputs further get fused by element-wise summation.

In detail, according to Eq.~\eqref{eq1}, the static branch can be formulated as:
 \begin{equation}
 {{\mathbf{Y}}_{static}} = \sum\limits_{k = 1}^K {\left( {{{\mathbf{G}}_{physical}} + {{\mathbf{G}}_{mask}}} \right)} {{\mathbf{X}}}{\mathbf{W}},
 \label{eq2}
 \end{equation}
where ${{\mathbf{G}}_{physical}}$ denotes the physical graph from the physical connections of human body. ${{\mathbf{G}}_{mask}}$ denotes the parameterized mask, which is used as an attention onto the physical graph. Following 2s-AGCN~\cite{shi2019two}, these two graphs are combined in an additive manner. ${{\mathbf{Y}}_{static}}$ is the output features of the static branch. The static branch extracts the static topology information of the skeleton data, which has been proven helpful for final prediction.

More importantly, the dynamic branch can be formulated as:
\begin{equation}
{{\mathbf{Y}}_{dynamic}} = {{\mathbf{G}}_{global}}{{\mathbf{X}}}{\mathbf{W'}},
\label{eq3}
\end{equation}
where ${{\mathbf{G}}_{global}}$ is the dynamic graph predicted by CeN. ${{\mathbf{Y}}_{global}}$ is the output of the dynamic branch, which extracts the global context-enriched topology of the skeleton data. Note that the learnable parameters $\mathbf{W'}$ are not shared between the static and dynamic branches.

After extracting the static and context-enriched topology features with the static branch and dynamic branch, a weighted summation operation is applied for fusion. That is:
\begin{equation}
{{\mathbf{Y}}} ={{\mathbf{Y}}_{dynamic}} +  \lambda{{\mathbf{Y}}_{static}},
 \label{eq4}
\end{equation}
where $\lambda$ is the weight used to balance ${{\mathbf{Y}}_{dynamic}}$ and ${{\mathbf{Y}}_{static}}$, as they may differ in magnitude.

After the topology features are aggregated, a TC-block is appended for temporal feature aggregation. A shortcut connection is added after the TC-block. As shown in Figure~\ref{fig1}, the complete Dynamic GCN framework is built by stacking 10 Dynamic GConv layers. Global average pooling layer and a fully-connected layer along with softmax are appended after the GConv layers for final classification. The numbers of output channels in the GConv layers are kept the same as ST-GCN~\cite{yanspatial}.

\subsubsection{Joint-level Feature Aggregation and Ensemble of Spatial-motion Modalities}

In the previous GCN-based methods~\cite{yanspatial,shi2019two,shi2019}, the number of nodes in the graph is kept unchanged. That is, given a skeleton sequence ${\bf{X}} \in {R^{C \times T \times N}}$, all GConv layers share the same number of joints $N$. In contrast, we propose a very simple way to gradually aggregate features at the joint level. Specifically, we use a projection matrix ${\bf{P}} \in {R^{N_i \times N_{i+1}}}$ to shrink the size of the joint dimension, where $N_{i+1} = \alpha N_i$ and $0 < \alpha  \leq 1$. We insert $\mathbf{P}$ into some intermediate layers of the graph convolutional network, so that $\widetilde {\mathbf{X}} = {\mathbf{X}} {\mathbf{P}} \in {R^{C \times T \times N_{i+1}}}$. By using the joint-level feature aggregation, the FLOPs of the model can be greatly reduced, and the performance of the model is barely affected.

Moreover, to further boost the performance, we explore multiple modalities, namely joint, bone and their corresponding motion modalities as Shi et al.~\cite{shi2019,shi2019two} did.
The temporal motion of skeleton joints has been shown informative to discriminate fine-grained actions, such as ``put on jacket'' and ``take off jacket''. For one person in frame $t$, each joint can be denoted as ${{\mathbf{J}}^t} \in {R^{3 \times N}}$. Then the joint motion $\mathbf{M}^t$ is defined as the temporal movement of each joint ${{\mathbf{M}}^t} = {{\mathbf{J}}^{t + 1}} - {{\mathbf{J}}^t}$.

Moreover, the bone is another spatial information, which has been proved important in previous works~\cite{shi2019two}. Each bone is defined as the vector pointing from a source joint to a target joint. It can be formulated as ${{\mathbf{B}}^t} = {\mathbf{J}}_{t\arg et}^t - {\mathbf{J}}_{source}^t$. Temporal motion can also be computed for the bone stream, in the same way as joint motion.

For ensemble of multiple modalities, we train one model per-modality separately. Then the logits before softmax of the four models are fused by summation.

\section{Experiments}
We evaluate Dynamic GCN on three large-scale skeleton-based action recognition benchmarks, namely NTU-RGB+D, NTU-RGB+D 120 and Skeleton-Kinetics. Extensive ablation studies are conducted to verify the impact of different components of the framework. Lastly, our final model is evaluated and compared with current state-of-the-arts.

\subsection{Datasets}

\noindent\textbf{NTU-RGB+D}~\cite{shahroudy} is the most widely used action recognition dataset. It contains 56,880 skeleton clips of 60 action classes. These clips were captured in the lab environment from three camera views. The annotations provide the 3D location $(x, y, z)$ of each joint in the camera coordinate system. There are 25 joints per-subject. Each clip is guaranteed to contain at most 2 subjects. We follow the standard evaluation protocols, namely cross-subject (C-Subject) and cross-view (C-View). In the C-subject setting, 40,320 clips from 20 subjects are used for training, and the rest for testing. In the C-View setting, 37,920 clips captured from camera 2 and 3 are used for training and those from camera 1 for testing.

\noindent\textbf{NTU-RGB+D 120}~\cite{liu2019ntu} is an extension of NTU-RGB+D, where the number of classes is expanded to 120 and the number of samples is expanded to 114,480. There are also two recommended evaluation protocols, namely cross-subject (C-Subject) and cross-setup (C-Setup). In the C-Subject setting, 63,026 clips from 53 subjects are used for training, and the remaining subjects are reserved for testing. In the C-Setup setting, 54,471 clips with even collection setup IDs are used for training, and the rest clips with odd setup IDs are used for testing.

\noindent\textbf{Skeleton-Kinetics} is derived from the Kinetics video action recognition dataset~\cite{kay2017kinetics}. The dataset contains 300,000 video clips of 400 classes. Each video clip lasts around 10 seconds. Human skeletons are estimated by \cite{yanspatial} from the RGB videos using the OpenPose toolbox~\cite{cao2018openpose}. Each joint consists of 2D coordinates $(x, y)$ in the pixel coordinate system and its confidence score $s$. Thus it is finally represented as a tuple of $(x, y, s)$. There are 18 joints for each person. In each frame at most two subjects are considered. We follow the same train-validation split as \cite{yanspatial}. That is, the training and validation sets contain 240,000 and 20,000 video clips respectively. Top-1 and top-5 accuracies are reported.

\subsection{Implementation Details}
Dynamic GCN is implemented in PyTorch~\cite{paszke2017automatic}. To verify the effectiveness of CeN on a high-performance baseline, we follow the same data processing strategy and the attention mechanism in MS-AAGCN~\cite{shi2019}. SGD~\cite{bottou2010large} with a 0.9 Nesterov momentum is used for optimization. The learning rate is set to 0.1 initially and reduced twice at the 35-th and 55-th epochs with a factor of 0.1. On all the three datasets, the model is trained for 65 epochs in total. And the $\lambda$ is set to 1. The joint aggregation rate $\alpha $ is set to 0.6, and the projection $\mathbf{P}$ is inserted twice after the 5-th and 8-th GConv layers. The input skeleton sequences are resized to a fixed length, i.e. 64 frames for both NTU-RGB+D and NTU-RGB+D 120, and 150 frames for Skeleton-Kinetics. For a fair comparison with methods based on ST-GCN, the number of spatial configurations is set to 3. To alleviate overfitting, weight decay is set to 0.0004. Batch size is set to 64, and cross-entropy loss is employed.

\subsection{Ablation Studies}

Ablation studies are conducted on the NTU-RGB+D dataset with the C-Subject setting. We first validate the effectiveness of the proposed CeN by comparing it with existing non-local-based methods. Next two alternative context modeling architectures are compared. Finally, the contribution of ensemble of spatial-motion modalities is reported.

\subsubsection{Choosing the Optimal Position of Joint Aggregation}

\begin{table}[t]
  \centering
  \caption{Impact of joint-level feature aggregation applied to different GConv layers.}
    \begin{tabular}{c|c|c|c|c|c}
    \toprule
    Position & 3\&6-th & 4\&7-th & 5\&8-th & 6\&9-th & w/o $\mathbf{P}$ \\
    \midrule
    Accuracy (\%) & 88.5  & 88.7  & \textbf{89.2} & 88.9  & 88.9 \\
    \bottomrule
    \end{tabular}%
  \label{tab:feat-agg}%
\end{table}%

For the joint-level feature aggregation with the projection matrix $\mathbf P$, the aggregation rate $\alpha$ is set to 0.6 empirically. To choose the appropriate GConv layers to apply the projection, we try a few configurations. As shown in Table~\ref{tab:feat-agg}, the position of applying $\mathbf P$ barely affects the performance. When applying it to the 5-th \& 8-th layers, it even slightly improves the accuracy from 88.9\% to 89.2\%.

\subsubsection{Effectiveness of CeN}


\begin{table}[t]
  \caption{Comparison of static, CeN-predicted and non-local-based graphs, as well as combinations of static and dynamic graphs. CeN predicts directed graphs, and CeN* predicts undirected graphs.}
  \label{table1}%
    \begin{tabular}{c|c|c|c|c}
    \toprule
    Static & Non-local & CeN*  & CeN   & Accuracy (\%)  \\
    \midrule
    $\checkmark$     &       &       &       & 88.2 \\
          & $\checkmark$     &       &       & 87.1 \\
          &       & $\checkmark$     &       & 88.1 \\
          &       &       & $\checkmark$     & \textbf{88.6} \\
    \midrule
    $\checkmark$     & $\checkmark$     &       &       & 88.4 \\
    $\checkmark$     &       & $\checkmark$     &       & 88.7 \\
    $\checkmark$     &       &       & $\checkmark$     & \textbf{89.2} \\
    \bottomrule
    \end{tabular}%
\end{table}%

As shown in Figure~\ref{fig4}, the Dynamic GConv layer is composed of static and dynamic branches. To measure the contribution of individual component, we train the model either using the static or dynamic branch only. To compare CeN with existing non-local-based method, we investigate the case when all CeNs are replaced with the non-local operation following~\cite{shi2019two}. Meanwhile, to verify the importance of directed graphs, we compare CeN with a variant CeN* which predicts undirected graphs by forcing the adjacency matrices to be symmetric.

As shown in Table~\ref{table1}, whether or not the static branch is enabled, the directed graph structure predicted by CeN achieves better performance than the undirected graph by CeN*. It clearly demonstrates the importance of directed graph, which can represent the dynamic characteristics of the skeleton more effectively. This also verifies our opinion that for different joints, different contextual information is preferred. That is why learning-based dynamic graph topology is necessary.

In addition, using static graph alone achieves an accuracy of 88.2\%, while non-local-based dynamic graph is inferior to static graph. CeN-predicted graph surpasses non-local-based graph by 1.5\%.

When combining static graph with dynamic graph, non-local-based graph barely improves the accuracy (i.e. by 0.2\%), while CeN-predicted graph significantly improves the accuracy from 88.2\% to 89.2\%, setting a new state-of-the-art for single model.

\subsubsection{Alternative Context-enriched Topology}

\begin{figure}[t]
  \begin{center}
\includegraphics[width=0.8\columnwidth]{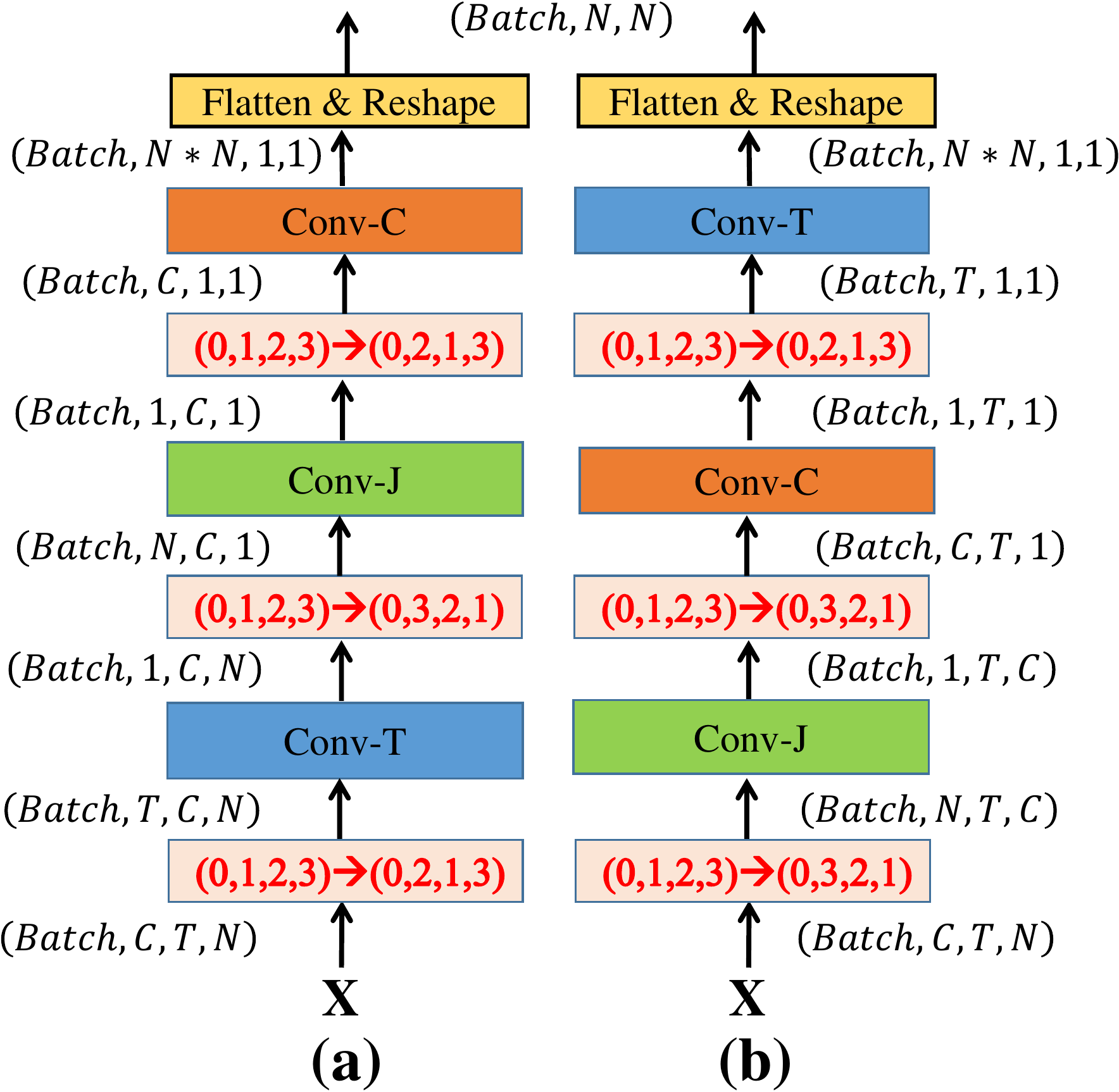} 
  \end{center}
\caption{Two variants of CeN. (a) CeN-F aggregates global context along the feature dimension. (b) CeN-T aggregates global context along the temporal dimension.}
\label{fig5}
\end{figure}


When learning the skeleton topology, CeN enriches contextual information globally along the joint dimension. Two straightforward alternatives are to learn contextual information along the feature or temporal dimension rather than the joint dimension. As shown in Figure~\ref{fig5}, they can be easily implemented by changing the order of the three $1\times 1$ convolutional layers. The two variants of CeN are referred to as CeN-F and CeN-T respectively.

\begin{table}[t]
  \caption{Comparison of CeN and two alternative context-encoding architectures. In all experiments the static branch is enabled. $\checkmark$ means that global context modeling is applied.}
  \label{table2}%
    \begin{tabular}{c|c|c|c|c}
    \toprule
    Methods  & Feature & Temporal  & Joint & Accuracy (\%) \\
    \midrule
    CeN-F   & $\checkmark$     &       &       & 88.5  \\
    CeN-T &       & $\checkmark$     &       & 88.4  \\
    CeN &       &       & $\checkmark$     & \textbf{89.2}  \\
    \bottomrule
    \end{tabular}%
\end{table}%

Table~\ref{table2} shows a comparison of CeN with its two variants. The performance of the model decreases by 0.7\% and 0.8\% for CeN-F and CeN-T respectively. The results verify our motivation that the surrounding joints are the most important for learning dynamic skeleton topology.

\subsubsection{Contribution of Model Ensemble}

\begin{table}[t]
  \caption{Comparison of spatial information, motion information and fusion of all modalities.}
  \label{table3}
    \begin{tabular}{c|c|c}
    \toprule
    Spatial Fusion & {Motion Fusion} & Accuracy (\%)  \\
    \midrule
    $\checkmark$     &       & 90.9 \\
          & $\checkmark$     & 87.1 \\
    $\checkmark$     & $\checkmark$     & \textbf{91.5} \\
    \bottomrule
    \end{tabular}%
\end{table}%

To fully exploiting the skeleton data, we evaluate the ensemble of four modalities, i.e. joints, bones, joint motion and bone motion. The modalities of joints and bones are combined as spatial information, and the modalities of joint motion and bone motion are combined as motion information. Using spatial and motion information alone as well as fusion of all modalities are evaluated.

As shown in Table~\ref{table3}, with ensemble of spatial modalities, the accuracy gets significantly improved from 89.2\% to 90.9\%. After combining the motion information, we achieve a top-1 accuracy of 91.5\% in the C-Subject setting of NTU-RGB+D, which is the new state-of-the-art for ensemble model.

\subsubsection{Comparison of FLOPs with Other GCN-based Methods}

In order to highlight the efficiency of the proposed Dynamic GCN, we compare it with existing GCN-based methods in terms of FLOPs and accuracy. For simplicity, for all methods, FLOPs and accuracy of single model are reported. As shown in Table~\ref{flops}, CeN only brings extra \textasciitilde7\% FLOPs for the baseline method, whose FLOPs increase from \textasciitilde1.86G to \textasciitilde1.99G. Compared with other methods, Dynamic GCN has $2\times$\textasciitilde$4\times$ advantage in terms of FLOPs, while achieving state-of-the-art performance.

\begin{table}[tbp]
  \centering
  \caption{Comparison of FLOPs and accuracy with other GCN-based methods.}
  \label{flops}
    \begin{tabular}{c|c|c}
    \toprule
      Method & FLOPs & \multicolumn{1}{c}{Accuracy (\%) }  \\
    \midrule
    ST-GCN~\cite{yanspatial} & \textasciitilde3.56G & 81.5   \\
    AS-GCN~\cite{li2019actional} & \textasciitilde6.10G  & 86.8   \\
    MS-AAGCN~\cite{shi2019} & \textasciitilde3.98G & 88.0     \\
    MS-G3D Net (1 pathway)~\cite{liu2020} & \textasciitilde5.21G & 89.1   \\
    MS-G3D Net (2 pathways)~\cite{liu2020} & \textasciitilde8.32G & \textbf{89.4}   \\
    \midrule
    Dynamic GCN (w/o CeN) & \textasciitilde1.86G & 88.2   \\
    Dynamic GCN (ours) & \textbf{\textasciitilde1.99G (+\textasciitilde7\%)} & 89.2    \\
    \bottomrule
    \end{tabular}%
\end{table}%

\subsubsection{Visualization of Learned Topology}
\begin{figure}[t]
\centering
\subfigure[Jump]{ \label{fig6:a}
\begin{minipage}[t]{0.33\linewidth}
\centering
\includegraphics[width=1in]{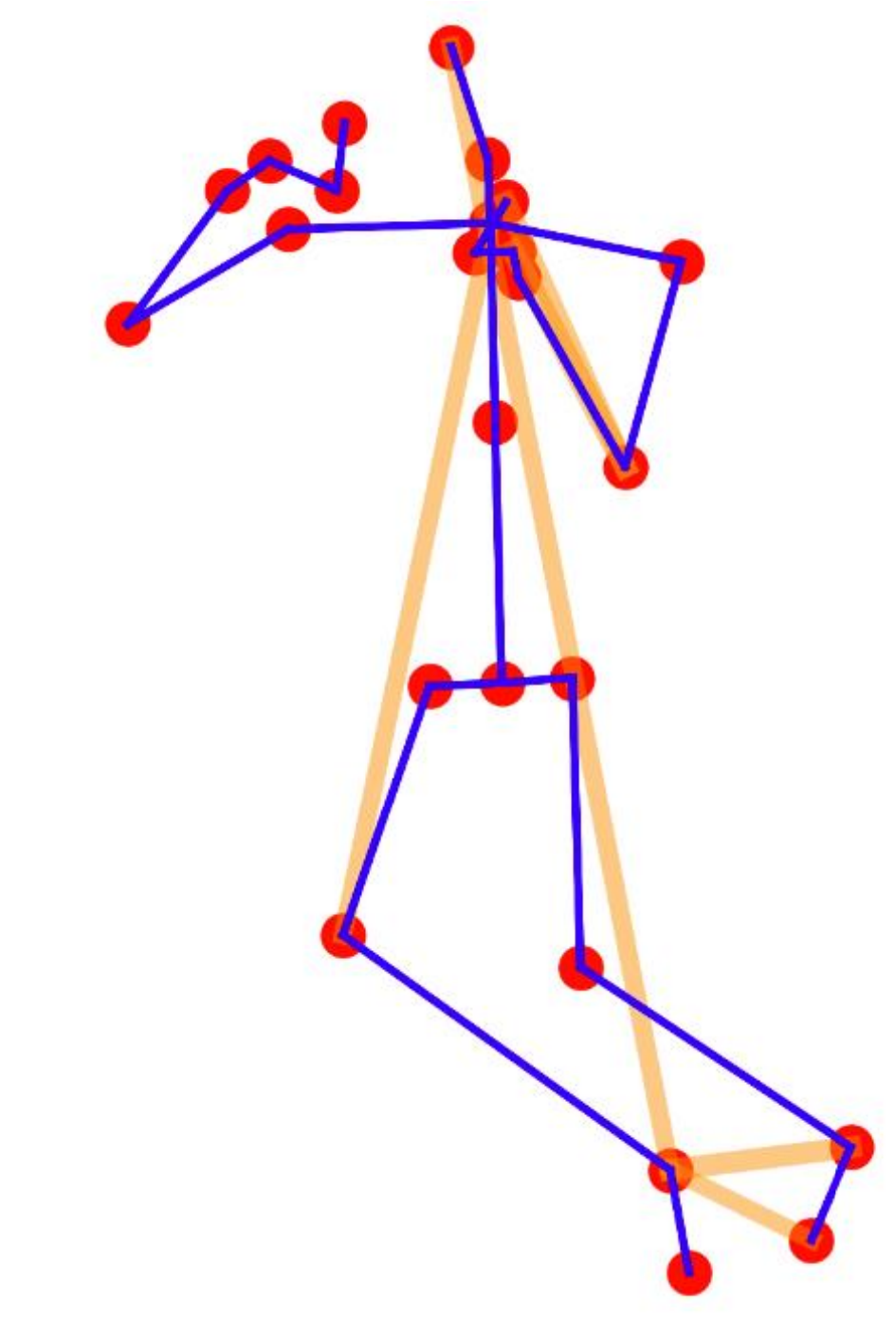}
\end{minipage}%
}%
\subfigure[Walking]{ \label{fig6:b}
\begin{minipage}[t]{0.33\linewidth}
\centering
\includegraphics[width=0.8in]{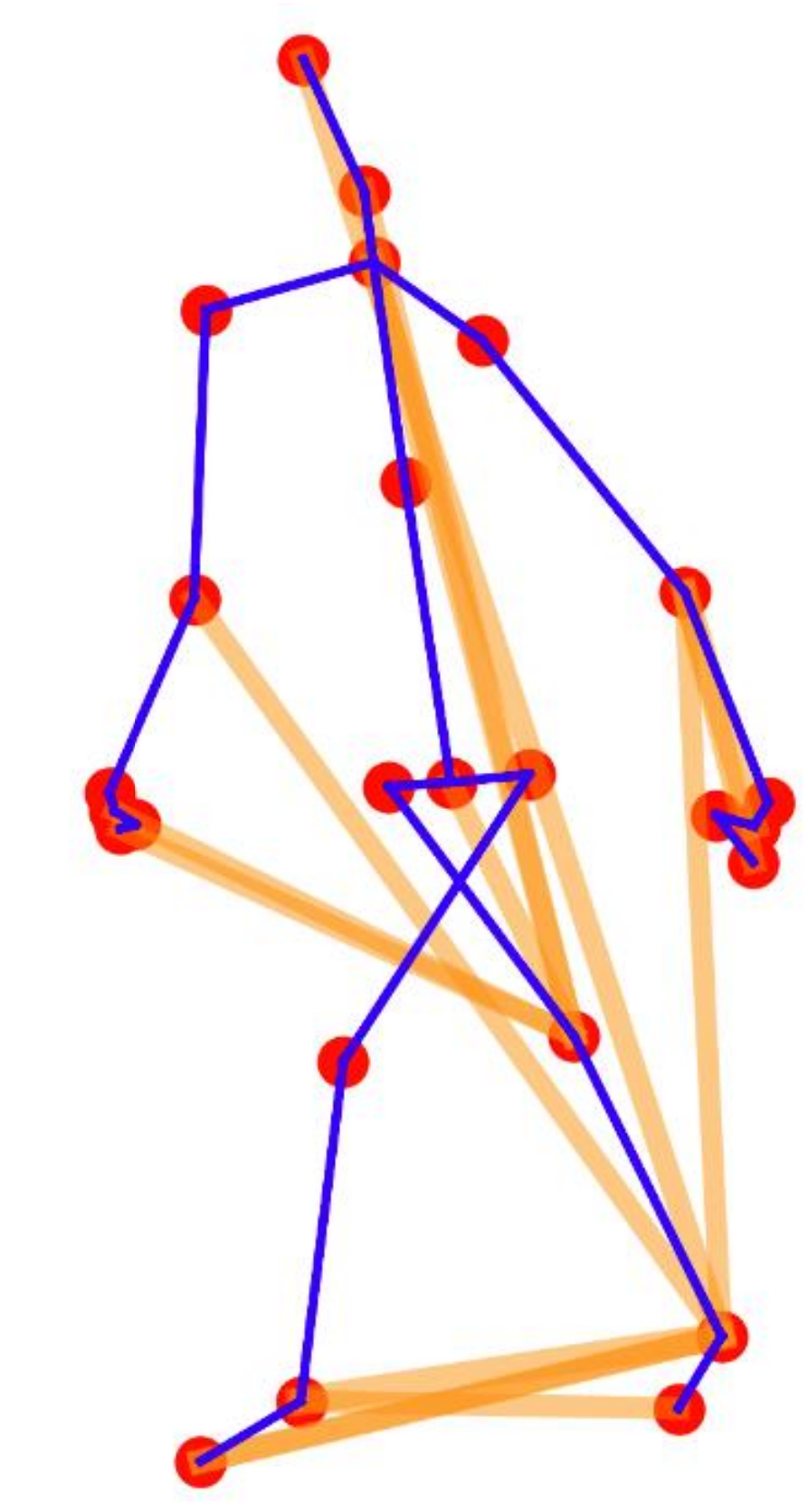}
\end{minipage}%
}%
\subfigure[Wipe Face]{ \label{fig6:c}
\begin{minipage}[t]{0.33\linewidth}
\centering
\includegraphics[width=0.8in]{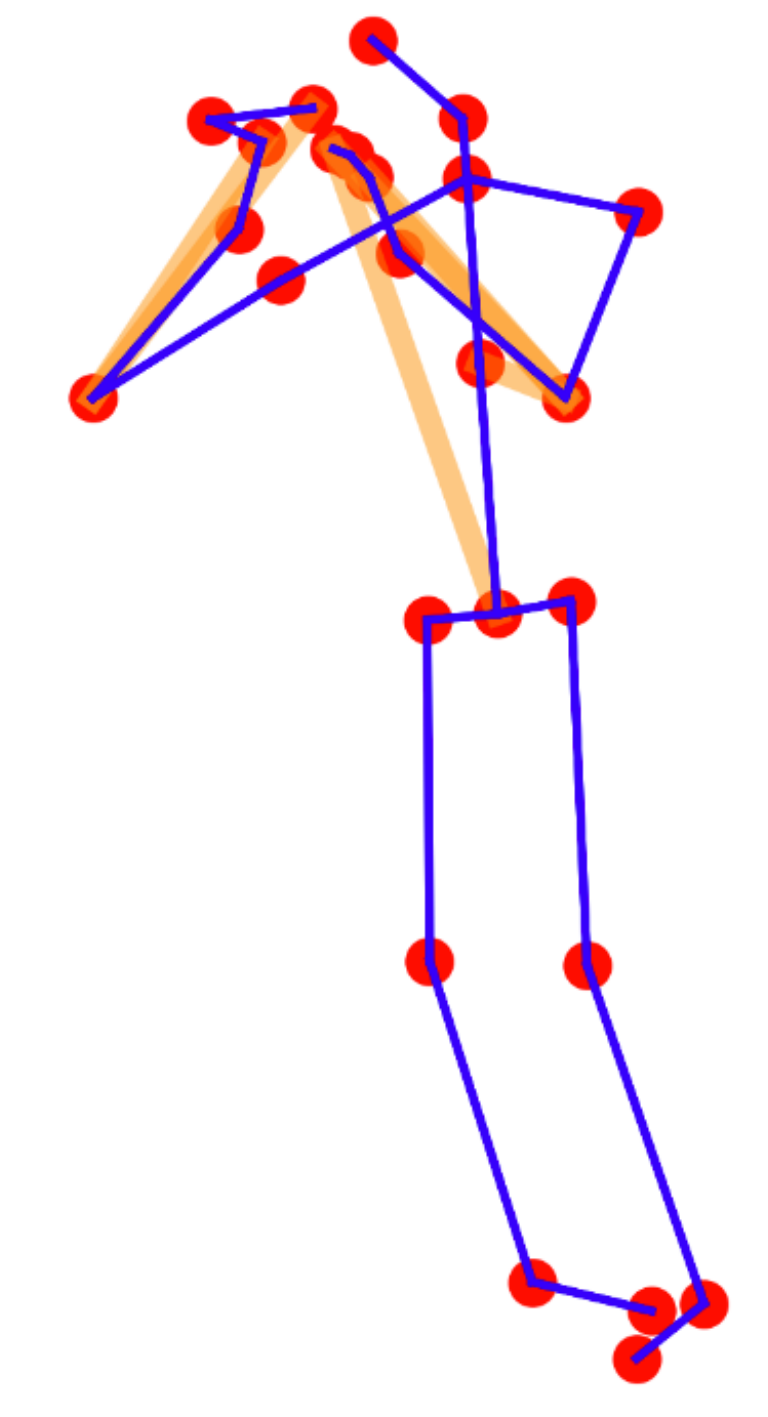}
\end{minipage}
}%
\centering
\caption{Visualization of learned topologies of three actions. Dynamic topologies (dark-orange lines) inferred by CeN are superimposed to physical topology (blue lines) of human body.}
\label{fig6}
\end{figure}

To investigate how well CeN learns the graph topology, we attempt to visualize the learned skeleton topology. Since CeN predicts dynamic graph for different samples, for clarification, we compute the average adjacency matrix of the 5-th GConv block for a specific action category. If the action involves two people, only the first person will be visualized.

Figure~\ref{fig6} visualizes the learned skeleton topologies of three actions. Response values larger than 0.4 are drawn as dark-orange lines. The blue lines are the physical connections of human body, and the red dots are the joints. As we can see, the strong dependencies captured by CeN are mainly related to two hands for the action of Wipe Face (Figure~\ref{fig6:c}) which is in line with cognition. For the action of Jump (Figure~\ref{fig6:a}), CeN mainly attends to the dependencies related to the knee joint and the feet joints. The action of walking (Figure~\ref{fig6:b}), CeN put attention to the hands and feet significantly. These results validate that reasonable potential topological patterns of different actions can be captured by CeN.

\subsection{Comparison with the State-of-the-arts}
We compare the proposed Dynamic GCN with other state-of-the-art methods on the NTU-RGB+D and Skeleton-Kinetics datasets in Table~\ref{table4} and Table~\ref{table5} respectively. For NTU-RGB+D 120, besides two recent methods, the baseline models mentioned in the original paper are also listed (Table~\ref{table6}).

\begin{table}[t]
  \caption{Performance comparison on the NTU-RGB+D dataset in both cross subject and cross view settings in terms of top-1 accuracy (\%).}
  \label{table4}
    \begin{tabular}{c|c|c|c}
    \toprule
    Method  & C-Subject & C-View & Year \\
    \midrule
    Lie Group~\cite{vemulapalli2014human}  & 50.1  & 82.8 & 2014  \\
    Deep LSTM~\cite{shahroudy} & 60.7  & 67.3& 2016 \\
    ST-LSTM+TS~\cite{liu2016spatio}  & 69.2  & 77.7& 2016 \\
    TCN~\cite{kim2017interpretable} & 74.3  & 83.1&2017  \\
    VA-LSTM~\cite{zhang2017view}  & 79.4  & 87.6& 2017 \\
    ST-GCN~\cite{yanspatial} & 81.5  & 88.3& 2018 \\
    DPRL~\cite{tang2018deep} & 83.5  & 89.8 &2018 \\
    HCN~\cite{lili} & 86.5  & 91.1&2018  \\
    GR-GCN~\cite{gao2019} & 87.5  & 94.3& 2019 \\
    AGC-LSTM~\cite{si2019attention} & 89.2  & 95.0& 2019 \\
    DGNN~\cite{shi2019skeleton}  & 89.9  & 96.1 &2019 \\
    STGR-GCN~\cite{li2019spatio} & 86.9  & 92.3 &2019 \\
    AS-GCN~\cite{li2019actional} & 86.8  & 94.2 &2019 \\
    2s-AGCN~\cite{shi2019two} & 88.5  & 95.1 &2019 \\
    MS-AAGCN~\cite{shi2019} & 90.0  & \textbf{96.2} & 2020\\
    MS-G3D Net~\cite{liu2020} & \textbf{91.5}  & \textbf{96.2}  &2020\\
    \midrule
    Dynamic GCN (ours) & \textbf{91.5 } & 96.0 &\\
    \bottomrule
    \end{tabular}%
\end{table}%

As shown in Table~\ref{table4}, Dynamic GCN achieves state-of-the-art performance, i.e. 91.5\% and 96.0\% in the C-Subject and C-View settings of NTU-RGB+D respectively. Although our accuracy is on par with MS-G3D Net~\cite{liu2020}, our model is much more efficient as analyzed in Table~\ref{flops}. \cite{yanspatial} and \cite{lili} are two representative methods for GCN-based and CNN-based methods. Dynamic GCN outperforms them by 10\% and 5\% respectively.
\begin{table}[t]
  \caption{Performance comparison on the Skeleton-Kinetics dataset. Both top-1 and top-5 accuracies (\%) are reported.}
  \label{table5}
    \begin{tabular}{c|c|c|c}
    \toprule
    Method & Top-1 Acc. & Top-5 Acc. & Year\\
    \midrule
    Deep LSTM~\cite{shahroudy} & 16.4  & 35.3 & 2016\\
    TCN~\cite{kim2017interpretable}   & 20.3  & 40.0 &2017\\
    ST-GCN~\cite{yanspatial} & 30.7  & 52.8&2018 \\
    STGR-GCN~\cite{li2019spatio} & 33.6  & 56.1& 2019 \\
    AS-GCN~\cite{li2019actional} & 34.8  & 56.5 & 2019 \\
    2s-AGCN~\cite{shi2019two} & 36.1  & 58.7 &2019\\
    MS-AAGCN~\cite{shi2019} & 37.8  & 61.0 &2020 \\
    MS-G3D Net~\cite{liu2020} & \textbf{38.0}  & 60.9 &2020 \\
    \midrule
    Dynamic GCN (ours) & 37.9 & \textbf{61.3} \\
    \bottomrule
    \end{tabular}%
\end{table}%

As for the Kinetics-Skeleton dataset (shown in Table~\ref{table5}), our model also achieves state-of-the-art performance (37.9\% top-1 accuracy and 61.3\% top-5 accuracy).

For the recently released NTU-RGB+D 120 dataset, so far few state-of-the-art methods have been evaluated on it. We compare the Dynamic GCN with the baseline methods mentioned in the original paper. As shown in Table~\ref{table6}, our model achieves a top-1 accuracy of 87.3\% and 88.6\% in the C-Subject and C-Setup settings respectively, surpassing the baseline methods by a large margin. Compared with MS-G3D Net~\cite{liu2020}, we achieve slightly better accuracy.

\begin{table}[t]
  \caption{Performance comparison on the NTU-RGB+D 120 dataset in both cross subject and cross setup settings in terms of top-1 accuracy (\%).}
  \label{table6}
    \begin{tabular}{c|c|c|c}
    \toprule
    Method  & C-Subject & C-Setup & Year\\
    \midrule
    Dynamic Skeleton~\cite{hu2015jointly} & 50.8  & 54.7 & 2015 \\
    PA-LSTM~\cite{shahroudy} & 25.5  & 26.3  & 2016\\
    ST-LSTM+TS~\cite{liu2016spatio} & 55.7  & 57.9 & 2016\\
    FSNet~\cite{liu2019skeleton} & 59.9  & 62.4  &2019\\
    2s-ALSTM~\cite{liu2017skeleton} & 61.2  & 63.3 &2017 \\
    MT CNN~\cite{ke2018learning} & 62.2  & 61.8 &2018 \\
    BPEM~\cite{liu2018recognizing}  & 64.6  & 66.9 &2018 \\
    2s-AGCN~\cite{shi2019two} & 82.9  & 84.9 &2019\\
    MS-G3D Net~\cite{liu2020} & 86.9  & 88.4 &2020\\
    \midrule
    Dynamic GCN (ours) & \textbf{87.3 } & \textbf{88.6 } \\
    \bottomrule
    \end{tabular}%
\end{table}%

\section{Discussion}
\subsection{On Hybrid GCN-CNN}
Compared with GCN alone, combining GCN and CNN into a hybrid GCN-CNN framework is indeed beneficial. In this work, the proposed Dynamic GCN is merely a simple form of the hybrid GCN-CNN architectures, in which CNN is used for topology learning. The capacity of GCN is greatly expanded with a flexible and expressive graph topology.

Beyond the task of skeleton-based action recognition, the hybrid GCN-CNN can be applied in many fields, such as social network modeling. Although the Dynamic GCN we propose is generally proven effective, the form of hybrid GCN-CNN requires further exploration for different tasks. For example, normal convolution and graph convolution can be stacked sequentially or in parallel for better skeleton feature learning. We will leave this as future work.

\subsection{On Learned Graph Toplogy}


For skeleton data, the physical graph enjoys the benefit of being definite and explainable. According to our experiment in Table~\ref{table1}, its performance is reasonable and competitive. Nevertheless, learned graph by CeN is complementary to physical graph, which indicates that there exist potential connections that are informative but missed by physical connections. Our visualization of learned topology in Figure~\ref{fig6} verifies this point. We believe that for other graph-structured data like social network, the proposed context-enriched topology learning is also applicable, considering the fact that the pre-defined topology is noisy, incomplete and unreliable.


\section{Conclusion}
How to extract the topology of skeleton data effectively for graph convolutional networks is the major challenge for skeleton-based action recognition. In this paper, we propose the Dynamic GCN framework, which leverages the advantages of GCN and CNN. The context-encoding network is designed to learn global context-enriched topology. Extensive experiments on three large-scale datasets validate the superiority of the proposed method. The significant reduction of FLOPs compared with existing methods makes our model more competitive for deployment, in particular on edge devices where computing power is limited. The novelly introduced hybrid GCN-CNN architecture as well as the technique of context-enriched topology learning may provide insights for future research on skeleton data analysis and beyond.


\bibliographystyle{ACM-Reference-Format}
\bibliography{main}


\end{document}